\documentstyle[aclap,gb4e,epsf]{article}

\title{An Annotation Scheme for Free Word Order Languages}

\author{Wojciech Skut, Brigitte Krenn, Thorsten Brants, Hans Uszkoreit\\
        Universit{\"a}t des Saarlandes\\
        66041 Saarbr{\"u}cken, Germany\\
        {\tt \{skut,krenn,brants,uszkoreit\}@coli.uni-sb.de}\\[1ex]
	{\em In Proceedings of the Fifth Conference on Applied Natural
	Language}\\ {\em Processing (ANLP), Washington, D.C., 1997}}

\begin{document}

\maketitle


\begin{abstract}

We describe an annotation scheme and a tool developed for  creating
linguistically annotated corpora for non-configurational languages.
Since the requirements for such a formalism differ from those posited
for configurational languages, several features have been added,
influencing the architecture of the scheme. The resulting scheme
reflects a stratificational notion of language, and makes only minimal
assumptions about the  interrelation of the particular representational
strata.

\end{abstract}


\section{Introduction}
\label{sec:intro}

The work reported in this paper aims at providing syntactically
annotated corpora (`treebanks') for stochastic grammar induction. In
particular, we focus on several methodological issues  concerning the
annotation of non-configurational  languages.

In section \ref{motiv}, we examine the appropriateness of existing
annotation schemes. On the basis of these considerations, we formulate
several additional requirements. A formalism complying with these 
requirements is described in section \ref{sec:form}. Section 
\ref{sec:treat} deals with the 
treatment of selected phenomena. For a
description of the annotation tool see section \ref{sec:tool}.


\section{Motivation}
\label{motiv}


\subsection{Linguistically Interpreted Corpora}

Combining raw language data with linguistic information offers a
promising basis for the development of new efficient and robust NLP
methods. Real-world texts annotated with different strata of linguistic
information can be used for grammar induction. The data-drivenness of
this approach presents a clear advantage over the traditional, idealised
notion of competence grammar.


\subsection{Existing Treebank Formats}

Corpora annotated with syntactic structures are commonly referred
to as {\em treebanks}. Existing treebank annotation schemes exhibit 
a fairly uniform architecture, as they all have to meet the same basic 
requirements, namely:

\begin{description}
\item[Descriptivity:]
	Grammatical phenomena are to be described rather than explained.
\item[Theory-independence:]
	Annotations should not be influenced by theory-specific
	considerations. Nevertheless, different theory-specific
	representations shall be recoverable from the annotation, cf.
	\cite{MarcusEA:94}.
\item[Multi-stratal representation:]
	Clear separation of different description levels is desirable.
\item[Data-drivenness:] 
	The scheme must provide representational means for all phenomena
	occurring in texts. Disambiguation is based on human processing
	skills (cf. \cite{MarcusEA:94}, \cite{Sampson:95}, \cite{BlackEA:96}).
\end{description}

The typical treebank architecture is as follows:

\begin{description}
\item[Structures:]
	A context-free backbone is augmented with trace-filler representations
	of non-local dependencies. The underlying {\em argument structure} is
	not represented directly, but can be recovered from the tree and
	trace-filler annotations.
\item[Syntactic category]
	is encoded in node labels.
\item[Grammatical functions]
	constitute a complex label system (cf. \cite{BiesEA:95},
	\cite{Sampson:95}).
\item[Part-of-Speech]
	is annotated at word level.
\end{description}

Thus the context-free constituent backbone plays a pivotal role in the
annotation scheme. Due to the substantial differences between existing
models of constituent structure, the question arises of how the {\em
theory independence} requirement can be satisfied. At this point the
importance of the underlying {\em argument structure} is emphasised (cf.
\cite{LehmannEA:96}, \cite{MarcusEA:94}, \cite{Sampson:95}).


\subsection{Language-Specific Features}
\label{sec:empir}

Treebanks of the format described in the above section have
been designed for English. Therefore, the solutions they offer
are not always optimal for other language types. As for free
word order languages, the following features may cause problems: 

\begin{itemize}
\item	local and non-local dependencies form a continuum rather
	than clear-cut classes of phenomena;

\item	there exists a rich inventory of discontinuous constituency
	types (topicalisation, scrambling, clause union, pied piping, 
	extraposition, split NPs and PPs);

\item	word order variation is sensitive to many factors, e.g. category,
	syntactic function, focus;

\item the grammaticality of different word permutations does not
	fit the traditional binary `right-wrong' pattern; it rather
	forms a gradual transition between the two poles.
\end{itemize}

In light of these facts, serious difficulties can be expected arising from
the structural component of the existing formalisms. Due to
the frequency of discontinuous constituents in non-configurational
languages, the filler-trace mechanism would be used very often, yielding
syntactic trees fairly different from the underlying predicate-argument
structures.

Consider the German sentence

\small
\begin{exe}
\ex \gll daran wird ihn Anna erkennen, da{\ss} er weint\\
        at-it will him Anna recognise that he cries\\
        `Anna will recognise him at his cry'
\end{exe}
\normalsize

A sample constituent structure is given below:

\begin{center}
\mbox{\epsffile{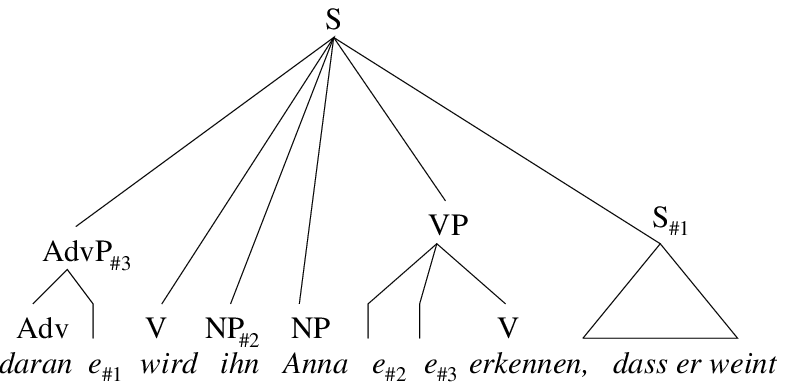}}
\end{center}

The fairly short sentence contains three non-local dependencies, marked
by co-references between traces and the corresponding nodes. This hybrid
representation makes the structure less transparent, and therefore more
difficult to annotate.

Apart from this rather technical problem, two further arguments speak
against phrase structure as the structural pivot of the annotation
scheme: 

\begin{itemize}
\item	Phrase structure models stipulated for non-configurational
	languages differ strongly from each other, presenting
	a challenge to the intended theory-independence of the scheme.
\item	Constituent structure serves as an explanatory device
	for word order variation, which is difficult
	to reconcile with the descriptivity requirement.
\end{itemize}

Finally, the structural handling of free word order means stating 
well-formedness constraints on structures involving many trace-filler 
dependencies, which has proved tedious. Since most methods of handling
discontinuous constituents make the formalism more powerful, the
efficiency of processing deteriorates, too.

An alternative solution is to make argument
structure the main structural component of the formalism. This
assumption underlies a growing number of recent syntactic theories which
give up the context-free constituent backbone, cf. \cite{McCawley:87},
\cite{Dowty:89}, \cite{Reape:93}, \cite{Kathol:Pollard:95b}. These
approaches provide an adequate explanation for several issues
problematic for phrase-structure grammars (clause union, extraposition,
diverse second-position phenomena).


\subsection{Annotating Argument Structure}

Argument structure can be
represented in terms of unordered trees (with crossing branches). In
order to reduce their ambiguity potential, rather simple, `flat' trees
should be employed, while more information can be expressed by a rich
system of function labels.

Furthermore, the required theory-independence means that the form of
syntactic trees should not reflect theory-specific assumptions, e.g.
every syntactic structure has a unique head. Thus,
notions such as {\em head} should be distinguished at the level of
syntactic functions rather than structures. This requirement speaks
against the traditional sort of {\em dependency trees}, in which heads
are represented as non-terminal nodes, cf. \cite{Hudson:wg}.

A tree meeting these requirements is given below:

\begin{center}
\mbox{\epsffile{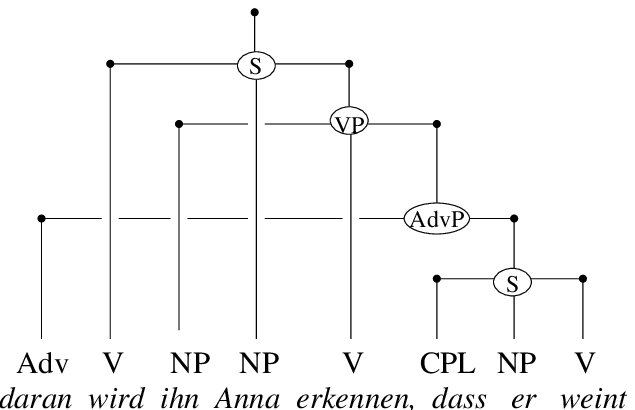}}
\end{center}

Such a word order independent representation has the advantage of all
structural information being encoded in a single data structure. A
uniform representation of local and non-local dependencies makes
the structure more transparent%
\footnote{A context-free constituent backbone can still be recovered from the
surface string and argument structure by reattaching `extracted' structures 
to a higher node.}.


\section{The Annotation Scheme}
\label{sec:form}


\subsection{Architecture}

We distinguish the following levels of representation:

\begin{description}
\item[Argument structure,]
	represented in terms of unordered trees.
\item[Grammatical functions,]
	encoded in edge labels, e.g. SB (subject), MO (modifier), HD
	(head).
\item[Syntactic categories,]
	expressed by category labels assigned to non-terminal nodes and by
	part-of-speech tags assigned to terminals.
\end{description}


\subsection{Argument Structure}

\begin{figure*}
\begin{center}
\epsfysize=51mm
\mbox{\epsffile{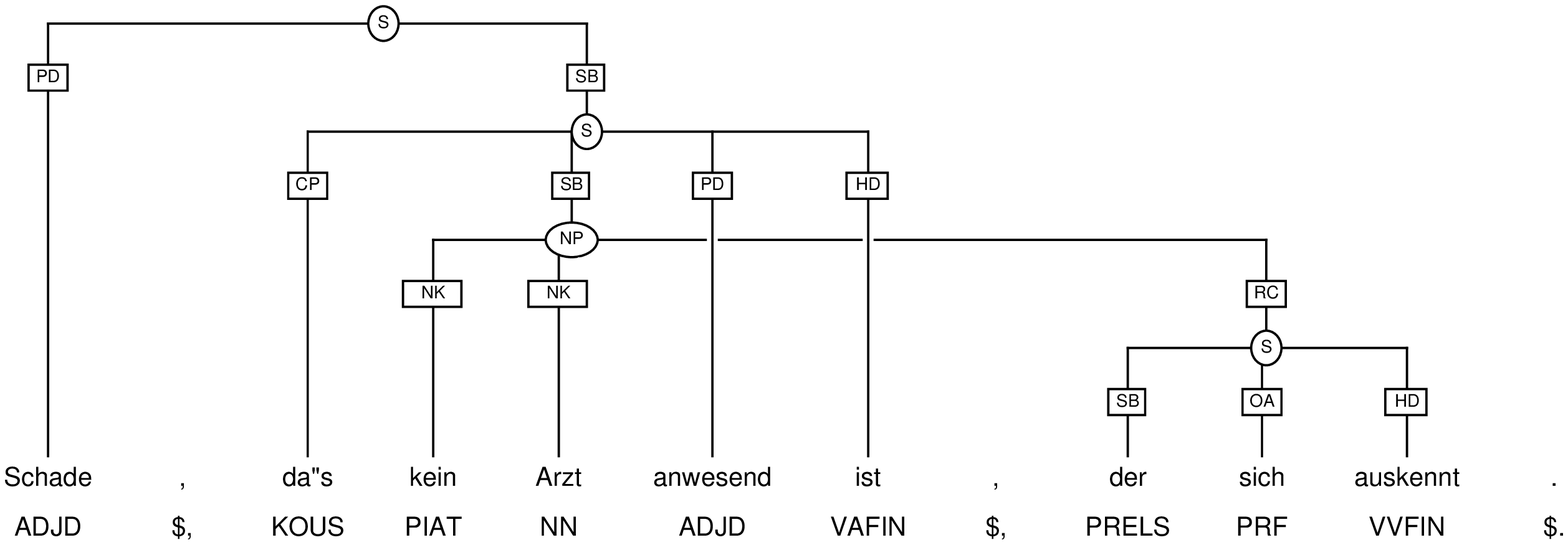}}
\vspace{-3mm}
\caption{Headed and non-headed structures, extraposition}
\label{fig:sms}

\bigskip
\smallskip

\epsfysize=44mm
\mbox{\epsffile{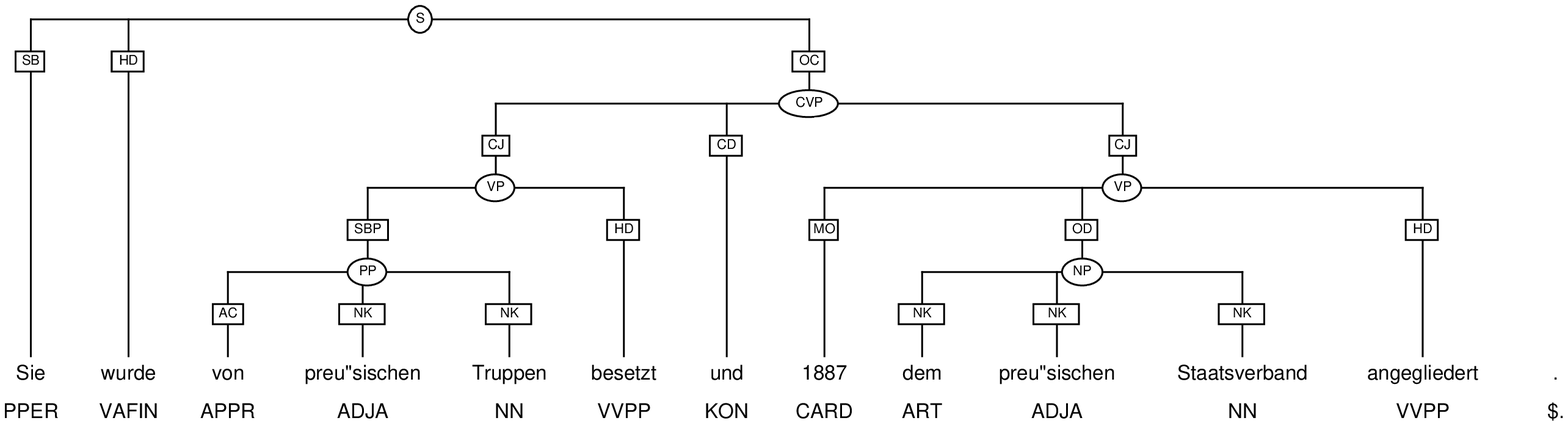}}
\vspace{-4mm}
\caption{Coordination}
\label{fig:coord}

\bigskip
\smallskip

\epsfysize=40mm
\mbox{\epsffile{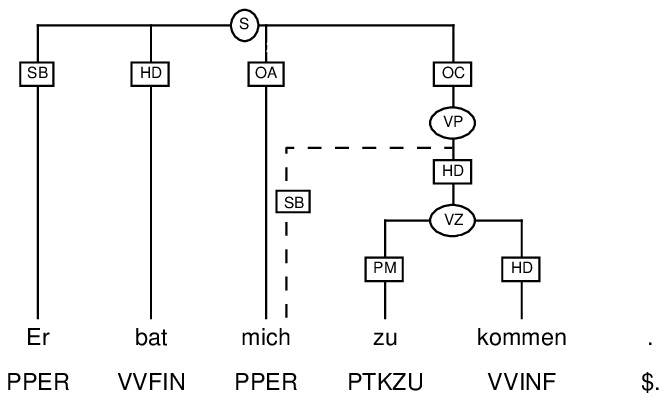}}
\vspace{-3mm}
\caption{Equi construction}
\label{fig:equi}

\end{center}
\end{figure*}

A structure for (\ref{sch}) is shown in fig. \ref{fig:sms}.

\small
\begin{exe}
\ex \gll schade, da{\ss} kein Arzt anwesend ist, der {sich auskennt}\\
         pity that no doctor present is who {is competent}\\
         `Pity that no competent doctor is here'
        \label{sch}
\end{exe}
\normalsize
Note that the root node does not have a head descendant (HD) as the
sentence is a predicative construction consisting of a subject (SB) and
a predicate (PD) without a copula. The subject is itself a sentence in
which the copula ({\em ist}) does occur and is assigned the tag HD%
\footnote{CP stands for complementizer, OA for accusative object and RC
for relative clause. NK denotes a `kernel NP' component (v. section
\ref{np}).}.

The tree resembles traditional constituent structures. The
difference is its word order independence: structural units
(``phrases'') need not be contiguous substrings. For instance,
the extraposed relative clause (RC) is  still treated as part 
of the subject NP.

As the annotation scheme does not distinguish different bar levels
or any similar intermediate categories, only a small set of node
 labels is needed (currently 16 tags, S, NP, AP \dots).


\subsection{Grammatical Functions}
\label{gf}

Due to the rudimentary character of the argument structure
representations, a great deal of
information has to be expressed by grammatical functions. Their further
classification must reflect different kinds of linguistic
information: morphology (e.g., case, inflection), category, dependency
type (complementation vs. modification), thematic role,
etc.\footnote{For an extensive use of grammatical functions cf.
\cite{KarlssonEA:95}, \cite{Voutilainen:94}.}

However, there is a trade-off between the granularity of information
encoded in the labels and the speed and accuracy of annotation. In order
to avoid inconsistencies, the corpus is annotated in two stages: {\em
basic annotation} and {\em refinement}. While in the first phase each
annotator has to annotate structures as well as categories and
functions, the refinement can be done separately for each representation
level.

During the first phase, the focus is on annotating correct structures
and a coarse-grained classification of grammatical functions, which
represent the following areas of information:

\begin{description}
\item[Dependency type:]
	{\em complements} are further classified according to features such as
	category and case: clausal  complements (OC), accusative objects (OA),
	datives (DA), etc. Modifiers are assigned the label MO (further
	classification with respect to thematic roles is planned). Separate
	labels are defined for dependencies that do not fit the
	complement/modifier dichotomy, e.g., pre- (GL) and postnominal
	genitives (GR).
\item[Headedness versus non-headedness:]
	Headed and non-headed structures are distinguished by the presence or
	absence of a branch labeled HD.
\item[Morphological information:]
	Another set of labels represents morphological information. PM stands
	for {\em morphological particle}, a label for German infinitival {\em
	zu} and superlative {\em am}. Separable verb prefixes are labeled
	SVP.
\end{description}

During the second annotation stage, the annotation is enriched with
information about thematic roles, quantifier scope and anaphoric
reference. As already mentioned, this is done separately for each of
the three information areas.


\subsection{Structure Sharing}

A phrase or a lexical item can perform multiple functions in a sentence.
Consider {\em equi verbs} where the subject of the infinitival VP is not
realised syntactically, but co-referent with the subject or object of
the matrix equi verb:

\small
\begin{exe}
\ex \gll er bat mich zu kommen\\
        he asked me to come \\ \label{equi}
\end{exe}
\normalsize
({\em mich} is the understood subject of {\em kommen}). In such cases,
an additional edge is drawn from the embedded VP node to the controller,
thus changing the syntactic tree into a graph. We call such additional
edges {\em secondary links} and represent them as dotted lines, see fig.
\ref{fig:equi}, showing the structure of (\ref{equi}).


\section{Treatment of Selected Phenomena}
\label{sec:treat}

As theory-independence is one of our objectives, the annotation scheme
incorporates a number of widely accepted linguistic analyses, especially
in the area of verbal, adverbial and adjectival syntax. However, 
some other {\em standard analyses} turn out to be problematic,
mainly due to the partial, idealised character of competence
grammars, which often marginalise or ignore such important phenomena as
`deficient' (e.g. headless) constructions, appositions, temporal
expressions, etc.

In the following paragraphs, we give annotations for a number of such 
phenomena.


\subsection{Noun Phrases}
\label{np}

Most linguistic theories treat NPs  as structures headed by a unique
lexical item (noun). However, this idealised model needs several
additional assumptions in order to account for such important phenomena
as complex nominal NP components (cf. (\ref{np1})) or nominalised
adjectives (cf. (\ref{np2})).

\small
\begin{exe}
\ex my uncle Peter Smith \label{np1}
\end{exe}

\begin{exe}
\ex \gll der sehr Gl{\"u}ckliche\\
        the very happy\\
        `the very happy one' \label{np2}
\end{exe}
\normalsize
In (\ref{np1}), different theories make different headedness
predictions. In (\ref{np2}), either a lexical nominalisation rule for
the adjective {\em Gl{\"u}ckliche} is stipulated, or the existence of an empty nominal head.
Moreover, the so-called DP analysis views the article {\em der} as the
head of the phrase. Further differences concern the attachment of the degree
modifier {\em sehr}.

Because of the intended theory-independence of the scheme, we annotate 
only the common minimum. We distinguish an NP kernel
consisting of determiners, adjective phrases and nouns. All components
of this kernel are assigned the label NK and treated as sibling nodes.

The difference between the particular NK's lies in the positional and
part-of-speech information, which is also sufficient to recover
theory-specific structures from our `underspecified' representations.
For instance, the first determiner among the NK's can be treated as the
specifier of the phrase. The head of the phrase can be determined in a
similar way according to theory-specific assumptions.

In addition, a number of clear-cut NP components can be defined outside
that juxtapositional kernel: pre- and postnominal genitives (GL, GR),
relative clauses (RC), clausal and sentential complements (OC). They are
all treated as siblings of NK's regardless of their position (in situ or
extraposed).


\subsection{Attachment Ambiguities}

Adjunct attachment often gives rise to structural ambiguities or
structural uncertainty. However, full or partial disambiguation takes
place in context, and the annotators do not consider unrealistic readings.

In addition, we have adopted a simple convention for those cases in
which context information is insufficient for total disambiguation: the
highest possible attachment site is chosen.

A similar convention has been adopted for constructions in which scope
ambiguities have syntactic effects but a one-to-one correspondence
between scope and attachment does not seem reasonable, cf. focus
particles such as {\em only} or {\em also}. If the scope of such a word
does not directly correspond to a tree node, the word is attached to the
lowest node dominating {\em all} subconstituents appearing in its scope.


\subsection{Coordination}

A problem for the rudimentary argument structure 
representations is the use of incomplete
structures in natural language, i.e. phenomena such as coordination and
ellipsis. Since a precise structural description of non-constituent
coordination would require a rich inventory of incomplete phrase types,
we have agreed on a sort of underspecified representations: the
coordinated units are assigned structures in which missing lexical
material is not represented at the level of primary links.  Fig.
\ref{fig:coord} shows the representation of the sentence:

\small
\begin{exe}
\ex \gll sie wurde von preu{\ss}ischen Truppen besetzt und 1887 dem preu{\ss}ischen Staat angegliedert\\
she was by Prussian troops occupied and 1887 to-the Prussian state incorporated\\
`it was occupied by Prussian troops and incorporated into Prussia in 1887'
\end{exe}
\normalsize
The category of the coordination is labeled CVP here, where C stands
for coordination, and VP for the actual category. This extra marking
makes it easy to distinguish between `normal' and coordinated
categories.

Multiple coordination as well as enumerations are annotated in the same
way. An explicit coordinating conjunction need not be present.

Structure-sharing is expressed using secondary links.


\section{The Annotation Tool}
\label{sec:tool}


\subsection{Requirements}

The development of linguistically interpreted corpora presents a
laborious and time-consuming task. In order to make the annotation
process more efficient, extra effort has been put into the development of
an annotation tool.

The tool supports immediate graphical feedback and automatic error
checking. Since our scheme permits crossing edges, visualisation as
bracketing and indentation would be insufficient. Instead, the complete
structure should be represented.

The tool should also permit a convenient handling of node and edge
labels. In particular, variable tagsets and label collections should be
allowed.


\subsection{Implementation}

As the need for certain functionalities becomes obvious with growing
annotation experience, we have decided to implement the tool in two
stages. In the first phase, the main functionality for building
and displaying unordered trees is supplied. In the second phase,
secondary links and additional structural functions are supported. The
implementation of the first phase as described in the following
para\-graphs is completed.

As keyboard input is more efficient than mouse input (cf.
\cite{LehmannEA:96}) most effort has been put in developing an efficient
keyboard interface. Menus are supported as a useful way of getting help
on commands and labels. In addition to pure annotation, we can attach
comments to structures.

\begin{figure*}[t]
\begin{center}
\hrule
\bigskip
\centerline{\epsffile{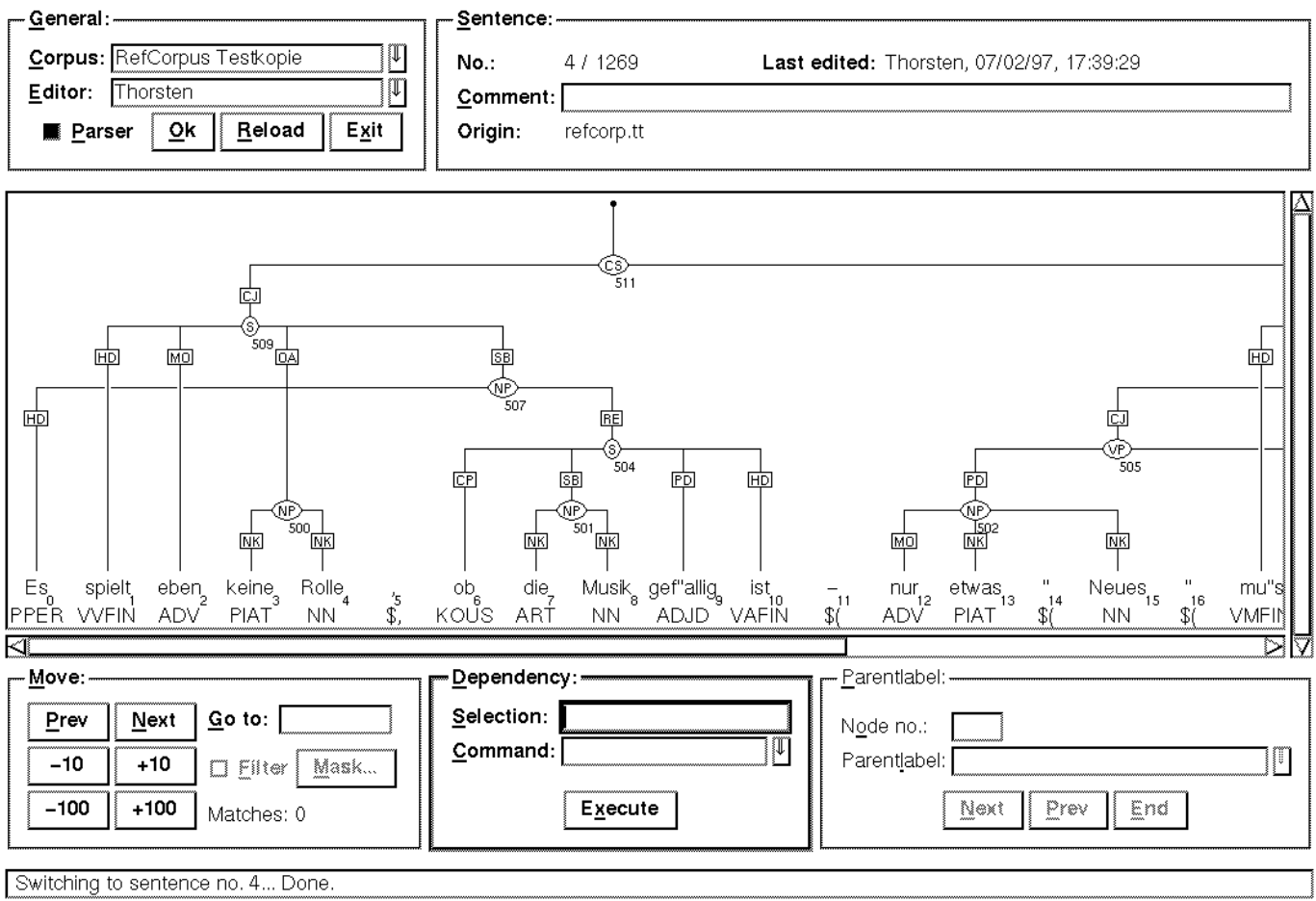}}
\bigskip
\hrule
\mbox{}
\caption{Screen dump of the annotation tool}
\label{FigScreen}
\end{center}
\end{figure*}

Figure \ref{FigScreen} shows a screen dump of the tool. The largest part
of the window contains the graphical representation of the structure
being annotated. The following commands are available:

\begin{itemize}
\item   group words and/or phrases to a new phrase;
\item   ungroup a phrase;
\item   change the name of a phrase or an edge;
\item   re-attach a node;
\item   generate the postscript output of a sentence.
\end{itemize}

The three tagsets used by the annotation tool (for words, phrases, and
edges) are variable and are stored together with the corpus. This allows
easy modification if needed. The tool checks the appropriateness of the
input.

For the implementation, we used Tcl/Tk Version 4.1. The corpus is stored
in a SQL database.


\subsection{Automation}

The degree of automation increases with the amount of data available.
Sentences annotated in previous steps are used as training material for
further processing. We distinguish five degrees of automation:

\begin{enumerate}
\item[0)] Completely manual annotation.
\item[1)] The user determines phrase boundaries and syntactic
	categories (S, NP, etc.). The program automatically assigns
	grammatical function labels. The annotator can alter the 
	assigned tags.
\item[2)] The user only determines the components of a new phrase, the
	program determines its syntactic category
	and the grammatical functions of its elements. Again, the
	annotator has the option of altering the assigned tags.
\item[3)] Additionally, the program performs simple bracketing,
	i.e., finds `kernel' phrases.
\item[4)] The tagger suggests partial or complete parses.
\end{enumerate}

So far, about 1100 sentences of our corpus have been annotated. This amount
of data suffices as training material to reliably assign the grammatical
functions if the user determines the elements of a phrase and its type
(step 1 of the list above).


\subsection{Assigning Grammatical Function Labels}

Grammatical functions are assigned using standard statistical 
part-of-speech tagging methods (cf. e.g. \cite{Cutting92} and 
\cite{Feldweg95}).

For a phrase $Q$ with children of type $T_1, \dots, T_k$ and grammatical
functions $G_1, \dots, G_k$, we use the lexical probabilities
\[
	P_Q(G_i|T_i)
\]
and the contextual (trigram) probabilities
\[
	P_Q(T_i|T_{i-1}, T_{i-2})
\]
The lexical and contextual probabilities are determined separately for
each type of phrase. During annotation, the highest rated grammatical
function labels $G_i$ are calculated using the Viterbi algorithm and
assigned to the structure, i.e., we calculate
{
\catcode`\_=8\catcode`\^=7
\[
	\mathop{\mbox{argmax}}\limits_G \prod_{i=1}^k P_Q(T_i|T_{i-1}, T_{i-2})
					\cdot P_Q(G_i|T_i).
\]
}

To keep the human annotator from missing errors made by the tagger, 
we additionally calculate the strongest competitor
for each label $G_i$. If its probability is close to the winner
(closeness is defined by a threshold on the quotient), the assignment is
regarded as unreliable, and the annotator is asked to confirm the assignment. 

For evaluation, the already annotated sentences were divided into two
disjoint sets, one for training (90\% of the corpus), the other
one for testing (10\%). The procedure was repeated 10 times with
different partitionings.

The tagger rates 90\% of all assignments as reliable and carries them out
fully automatically. Accuracy for these cases is 97\%. Most errors 
are due to wrong identification of the subject and
different kinds of objects in sentences and VPs. Accuracy of the
unreliable 10\% of assignments is 75\%, i.e., the annotator has to alter
the choice in 1 of 4 cases when asked for confirmation. Overall
accuracy of the tagger is 95\%.

Owing to the partial automation, the average annotation efficiency
improves by 25\% (from around 4 minutes to 3 minutes per sentence).

\section{Conclusion}

As the annotation scheme described in this paper focusses on annotating
argument structure rather than constituent trees, it differs from
existing treebanks in several aspects. These differences can be
illustrated by a comparison with the Penn Treebank annotation scheme.
The following features of our formalism are then of particular
importance:

\begin{itemize}
\item	simpler (i.e. `flat') representation structures
\item	complete absence of empty categories
\item	no special mechanisms for handling discontinuous constituency 
\end{itemize}

The current tagset comprises only 16 node labels and 34 function
tags, yet a finely grained classification will take place in the 
near future.

We have argued that the selected approach is better suited for 
producing high quality interpreted corpora in languages exhibiting free
constituent order. In general, the resulting interpreted data also are
closer to semantic annotation and more neutral with respect to
particular syntactic theories. 

As modern linguistics is also becoming more aware of the importance of
larger sets of naturally occurring data, interpreted corpora are a
valuable resource for theoretical and descriptive linguistic research. 
In addition the approach provides empirical material for
psycholinguistic investigation, since preferences for the choice of
certain syntactic constructions, linearizations, and attachments that
have been observed in online experiments of language production and
comprehension can now be put in relation with the frequency of these
alternatives in larger amounts of texts.  

Syntactically annotated corpora of German have been missing until now. 
In the second phase of the project Verbmobil a treebank for 30,000
German spoken sentences as well as for the same amount of English and
Japanese sentences will be created.  We will closely coordinate the
further development of our corpus with the annotation work in Verbmobil
and with other German efforts in corpus annotation. 

Since the combinatorics of syntactic constructions creates a demand for
very large corpora, efficiency of annotation is an important criterion
for the success of the developed methodology and tools. Our annotation
tool supplies efficient manipulation and immediate visualization of
argument structures. Partial automation included in the current version
significantly reduces the manual effort. Its extension is subject to
further investigations.


\section{Acknowledgements}

This work is part of the DFG Sonderforschungsbereich 378 {\em
Resource-Adaptive Cognitive Processes}, Project C3 {\em Concurrent
Grammar Processing}.

We wish to thank Tania Avgustinova, Berthold Crysmann, Lars Konieczny,
Stephan Oepen, Karel Oliva, Christian Wei{\ss}  and two anonymous
reviewers for their helpful comments on the content of this paper. We
also wish to thank Robert MacIntyre and Ann Taylor for valuable
discussions on the Penn Treebank annotation. Special thanks go to Oliver
Plaehn, who implemented the annotation tool, and to our fearless
annotators Roland Hendriks, Kerstin Kl\"ockner, Thomas Schulz, and
Bernd-Paul Simon.


\bibliographystyle{fullname}

\begin{thebibliography}{}
\bibitem[\protect\citename{Bies \bgroup et al.\egroup }1995]{BiesEA:95}
Ann~Bies et~al.
\newblock 1995.
\newblock {\em Bracketing Guidelines for Treebank II Style Penn Treebank Project}.
\newblock Technical report, University of Pennsylvania.

\bibitem[\protect\citename{Black \bgroup et al. \egroup }1996]{BlackEA:96}
Ezra~Black et~al.
\newblock 1996.
\newblock {Beyond Skeleton Parsing: Producing a Comprehensive Large-Scale
  General-English Treebank With Full Grammatical Analysis}.
\newblock In {\em The 16th International Conference on Computational
  Linguistics}, pages 107 -- 113, Copenhagen, Denmark.

\bibitem[\protect\citename{Cutting \bgroup et al.\egroup }1992]{Cutting92}
Doug Cutting, Julian Kupiec, Jan Pedersen, and Penelope Sibun.
\newblock 1992.
\newblock A practical part-of-speech tagger.
\newblock In {\em Proceedings of the 3rd Conference on Applied Natural Language
  Processing (ACL)}, pages 133--140.

\bibitem[\protect\citename{Dowty}1989]{Dowty:89}
David Dowty.
\newblock 1989.
\newblock Towards a minimalist theory of syntactic structure.
\newblock In {\em Tilburg Conference on Discontinuous Constituency}.

\bibitem[\protect\citename{Feldweg}1995]{Feldweg95}
Helmut Feldweg.
\newblock 1995.
\newblock Implementation and evaluation of a German HMM for POS disambiguation.
\newblock In {\em Proceedings of EACL-SIGDAT-95 Workshop}, Dublin, Ireland.

\bibitem[\protect\citename{Hudson}1984]{Hudson:wg}
Richard Hudson.
\newblock 1984.
\newblock {\em Word Grammar}.
\newblock Basil Blackwell Ltd.

\bibitem[\protect\citename{Karlsson \bgroup et al.\egroup }1995]{KarlssonEA:95}
Fred Karlsson, Atro Voutilainen, Juha Heikkila, and Arto Anttila.
\newblock 1995.
\newblock {\em Constraint Grammar. A Language-Independent System for Parsing
  Unrestricted Text}.
\newblock Mouton de Gruyter, Berlin, New York.


\bibitem[\protect\citename{Kathol and Pollard}1995]{Kathol:Pollard:95b}
Kathol, Andreas and Carl Pollard.
\newblock 1995.
\newblock {Extraposition via Complex Domain Formation}.
\newblock In {\em Proceedings of the $33^{rd}$ Annual Meeting of the ACL},
  pages 174--180, Cambridge, MA. Association for Computational Linguistics.

\bibitem[\protect\citename{Lehmann \bgroup et al.\egroup}1996]{LehmannEA:96}
Sabine~Lehmann et~al.
\newblock 1996.
\newblock {TSNLP -- Test Suites for Natural Language Processing}.
\newblock In {\em The 16th International Conference on Computational
  Linguistics}, pages 711 -- 717, Copenhagen, Denmark.

\bibitem[\protect\citename{Marcus \bgroup et al.\egroup } 1994]{MarcusEA:94}
Mitchell~Marcus et~al.
\newblock 1994.
\newblock The Penn Treebank: Annotating Predicate Argument Structure.
\newblock In {\em Proceedings of the Human Language Technology Workshop}, San
  Francisco. Morgan Kaufmann.

\bibitem[\protect\citename{McCawley} 1987]{McCawley:87}
James McCawley.
\newblock 1987.
\newblock Some additional evidence for discontinuity.
\newblock In Huck and Ojeda (eds.), {\em Discontinuous Constituency:
	Syntax and Semantics}, pp 185--200. New York, Academic Press.

\bibitem[\protect\citename{Reape}1993]{Reape:93}
Mike Reape.
\newblock 1993.
\newblock {\em A Formal Theory of Word Order: A Case Study in West Germanic}.
\newblock {Ph.D.} thesis, University of Edinburgh.

\bibitem[\protect\citename{Sampson}1995]{Sampson:95}
Geoffrey Sampson.
\newblock 1995.
\newblock {\em English for the Computer. The SUSANNE Corpus and Analytic
  Scheme}.
\newblock Clarendon Press, Oxford.

\bibitem[\protect\citename{Voutilainen}1994]{Voutilainen:94}
Atro Voutilainen.
\newblock 1994.
\newblock {\em Designing a Parsing Grammar}.
\newblock University of Helsinki, Dept. of General Linguistics.
Publications No. 22.
\end{thebibliography}


\clearpage

\end{document}